\documentclass[conference]{IEEEtran}
\IEEEoverridecommandlockouts
\usepackage{cite}
\usepackage{amsmath,amssymb,amsfonts}
\usepackage{algorithmic}
\usepackage{graphicx}
\usepackage{textcomp}
\usepackage{multirow}
\usepackage{xcolor}
\usepackage{mathrsfs} % 花体符号
\usepackage{amsmath}
\usepackage[misc]{ifsym}
\usepackage[bottom]{footmisc}

\def\BibTeX{{\rm B\kern-.05em{\sc i\kern-.025em b}\kern-.08em
    T\kern-.1667em\lower.7ex\hbox{E}\kern-.125emX}}
\begin{document}

\title{Distillation-Enhanced Physical Adversarial Attacks}


\author{
\begin{tabular}{ccc}
\textbf{Wei Liu$^*$} & \textbf{Yonglin Wu$^*$} & \textbf{Chaoqun Li} \\
Tsinghua University & Tsinghua University & Tsinghua University \\
Beijing, China & Beijing, China & Beijing, China \\
\small{tn18810525331@gmail.com} & \small{shadowuyl@foxmail.com} & \small{lichaoqunuestc@gmail.com} \\
\end{tabular}
& & \\
& & \\
\begin{tabular}{cc}
\textbf{Zhuodong Liu} & \textbf{Huanqian Yan \Letter} \\
Tsinghua University & Beihang University \\
Beijing, China & Beijing, China \\
\small{lzd19981105@gmail.com} & \small{yanhq@buaa.edu.cn} \\
\end{tabular}
}
%\thanks{$^*$ The authors contributed equally to this work.}
% \author{
% \textbf{Wei Liu$^{1}$}, \textbf{Yonglin Wu$^{1}$}, \textbf{Chaoqun Li$^{1}$}, \textbf{Zhuodong Liu$^{1}$}, \textbf{Huanqian Yan$^{2}$} \\
% $^{1}$ Tsinghua University, Beijing, China \\
% $^{2}$ Beihang University, Beijing, China \\
% \small{\{tn18810525331, lichaoqunuestc, lzd19981105\}@gmail.com, shadowuyl@foxmail.com, yanhq@buaa.edu.cn}
% }

\maketitle

\begin{abstract}
The study of physical adversarial patches is crucial for identifying vulnerabilities in AI-based recognition systems and developing more robust deep learning models. While recent research has focused on improving patch stealthiness for greater practical applicability, achieving an effective balance between stealth and attack performance remains a significant challenge. To address this issue, we propose a novel physical adversarial attack method that leverages knowledge distillation. Specifically, we first define a stealthy color space tailored to the target environment to ensure smooth blending. Then, we optimize an adversarial patch in an unconstrained color space, which serves as the ``teacher" patch. Finally, we use an adversarial knowledge distillation module to transfer the teacher patch’s knowledge to the ``student" patch, guiding the optimization of the stealthy patch. Experimental results show that our approach improves attack performance by 20\%, while maintaining stealth, highlighting its practical value.
\end{abstract}


\section{Introduction}
\label{sec:intro}
Adversarial attacks pose significant threats to deep learning-based applications, such as facial recognition~\cite{he2024mysticmask} and autonomous driving~\cite{deng2020analysis,ding2024invisible}. Studies on adversarial attacks can help uncover the mechanisms underlying model vulnerabilities, further enhance model robustness, and aid in preventing malicious attacks and protecting privacy. Adversarial attacks can be broadly classified into digital and physical attacks. Compared to digital attacks, physical attacks present a greater threat, as they can directly impact real-world objects. 

Adversarial patches are a common form of physical attack used to deceive deep learning models by applying patterns directly onto objects' surfaces. Obviously, the adversarial patch without color limit would be easily noticed by the human eye because it is not coordinated with the environmental background. Therefore, adversarial stealthiness is one of the key factors for adversarial patch applications.

%Recent works have focused on enhancing the stealthiness of these patches to improve their practicality.
UPC~\cite{huang2020universal} introduces an optimization constraint to generate patterns that appear natural to human observers. Building on this, NAP~\cite{hu2021naturalistic} leverages adversarial generative networks to create patches resembling animals, such as dogs. In contrast, DAP~\cite{guesmi2024dap} incorporates a novel objective function with a similarity metric to refine the patch generation process further. Some research has attempted to improve the stealthiness of adversarial patches by limiting the color space, resulting in color-constrained adversarial patches~\cite{hu2023physically,li2024prompt,li2024capgen}. While these methods achieve better stealthiness, they often come at the cost of reduced attack effectiveness, diminishing their practicality in real-world applications. Striking a balance between effectiveness and stealthiness remains a significant challenge, as improving one often compromises the other.
\begin{figure}[tb]
    \centering
    \includegraphics[width=0.95\columnwidth]{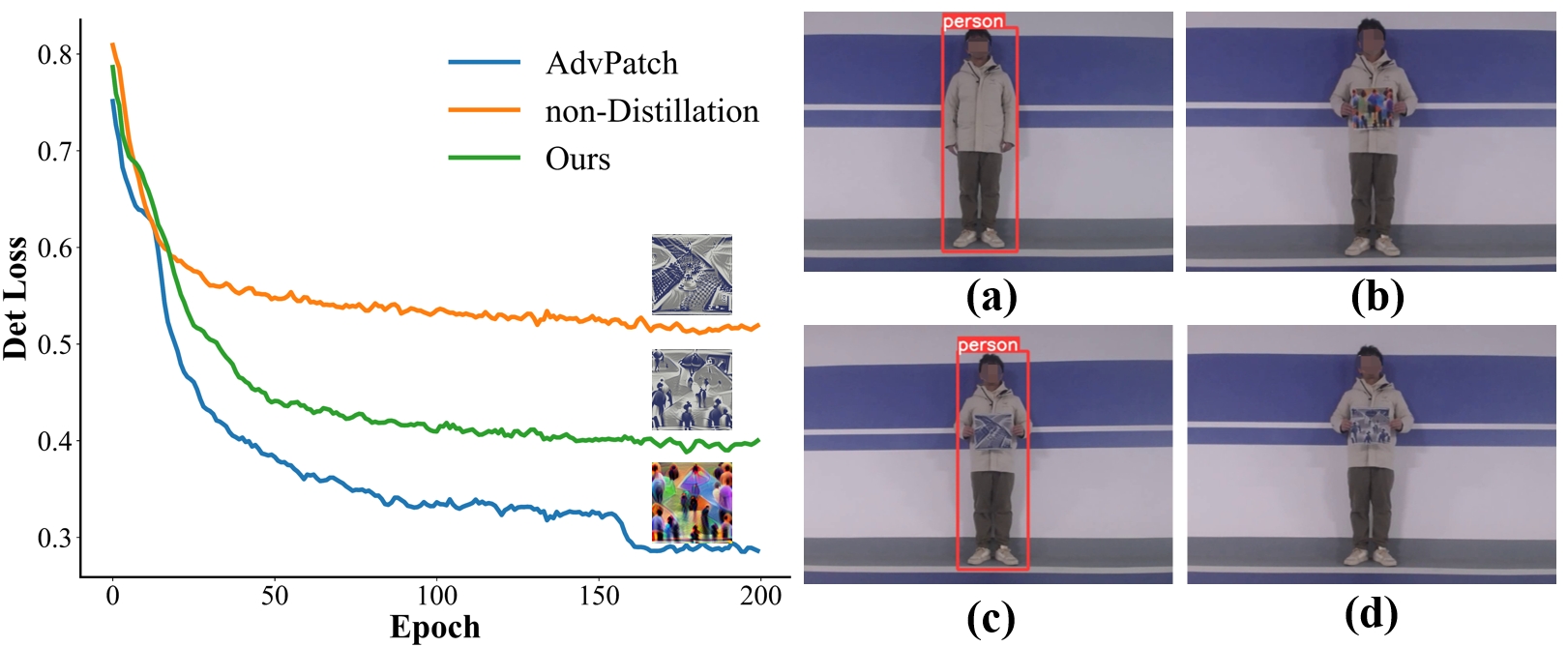}
    \caption{Left: The training detection box loss. A significant gap exists between the non-distillation and AdvPatch~\cite{thys2019fooling} methods. Our distillation-based method significantly narrows this gap. Right: Adversarial patches generated through various methods. (a) No adversarial patch. (b) AdvPatch. (c) non-Distillation. (d) Ours (Distillation). Notably, our distillation-based method enhances attack performance while preserving the same level of environmental concealment as the non-distilled method.}
    \label{pic:intro}
\end{figure}

\begin{figure*}[t]
    \centering
    \includegraphics[width=0.95\textwidth]{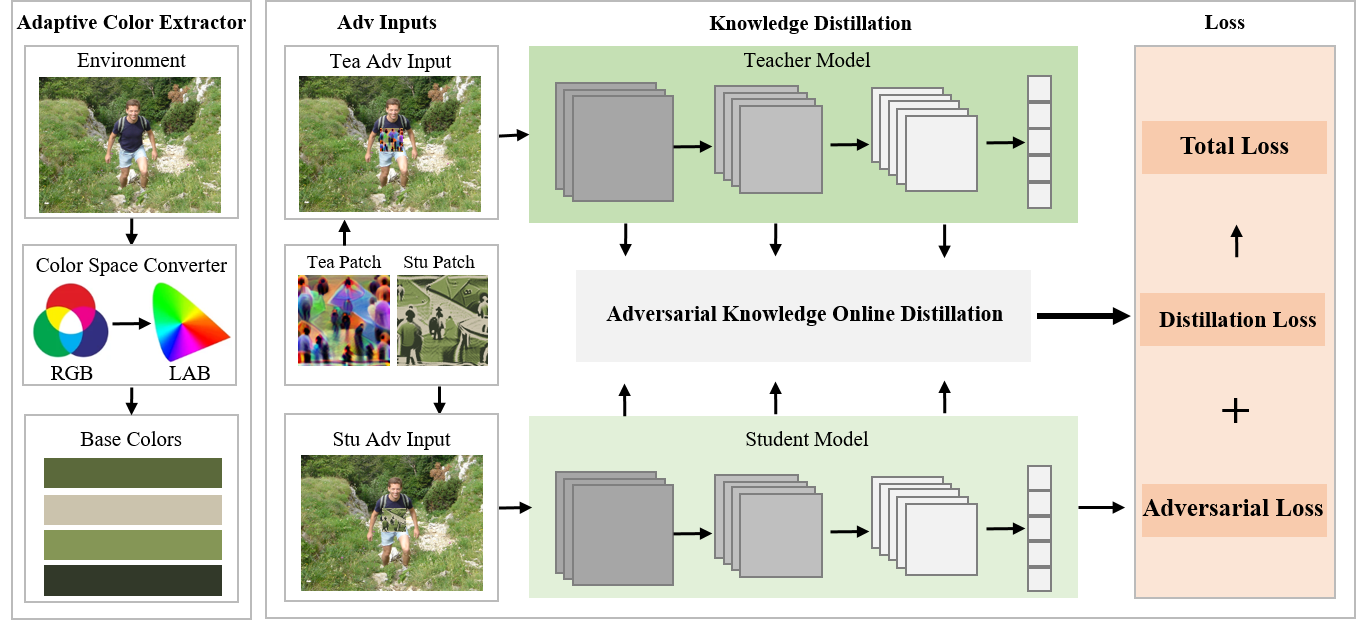}
    \caption{
    The overview of our proposed method. First, we extract the base colors from the environment to craft an adversarial patch that blends seamlessly with the environment. Next, we leverage a knowledge distillation approach, using a color-unconstrained adversarial patch to guide the generation of the stealthy patch, thereby enhancing its attack effectiveness.
    }
    \label{pic:framework}
\end{figure*}
To address this issue, a novel approach leveraging knowledge distillation is proposed in this paper. Inspired by the saliency mechanism of the visual cortex~\cite{zhaoping2014v1,zhaoping2008attention}, which highlights the human eye's sensitivity to inconsistent regions, we constrain the optimization of adversarial patches to a stealthy color space that blends with the surrounding environment, ensuring inconspicuousness. However, this constraint on the optimization space reduces the patch’s effectiveness. To mitigate this limitation, we introduce a knowledge distillation framework where a color-unconstrained adversarial patch acts as a teacher model, guiding a student model to generate patches within the stealthy color space. Through this framework, the student model learns adaptive adversarial features from the teacher, effectively enhancing attack performance while maintaining stealthiness. Some comparison results of the proposed method are shown in Fig.~\ref{pic:intro}, and the pipeline of the proposed method is shown in Fig.~\ref{pic:framework}. The main contributions can be summarized as follows:

\begin{itemize}

\item To improve the adversarial stealthiness, we have proposed to optimize adversarial texture with stealthy colors derived from the dominant colors of the environment.

\item 
We propose a knowledge distillation-based framework for generating adversarial patches to balance their attack performance and stealthiness. To the best of our knowledge, we are the first to apply knowledge distillation to enhance the performance of adversarial patches.

\item  Experimental results across multiple classic detection models demonstrate that our method improves attack performance by over 20\% compared to non-distillation methods. It indicates that our method can significantly enhance the adversarial attack while maintaining stealthiness.
\end{itemize}

\section{Related Work}
\label{sec:related_work}
\subsection{Physical Adversarial Attack}
Physical adversarial attacks pose a direct threat to systems such as stop sign recognition~\cite{song2018physical}, facial recognition~\cite{sharif2016accessorize} and pedestrian detection~\cite{thys2019fooling}. Among these, adversarial patches~\cite{song2018physical,zhao2019seeing,thys2019fooling} and adversarial texture onto clothing~\cite{xu2020adversarial} are particularly prevalent. However, adversarial patches or textures are often easily detectable due to their prominent contrast in color with the surrounding environment. To solve this problem, recent works has focused on enhancing their inconspicuousness. IAP~\cite{bai2021inconspicuous} leverages gradient-based interpretation mechanisms to decide the patch locations and employs a multi-scale generative adversarial network to produce adversarial patches that closely resemble the original image. Unlike generating common patterns, such as dogs~\cite{hu2021naturalistic,guesmi2024dap}, AdvCaT~\cite{hu2023physically} employs Voronoi diagrams and the Gumbel-softmax trick to parameterize camouflage textures, making them less noticeable. While these methods appear natural, they still lack true stealthiness. Other methods, such as leveraging natural light~\cite{zhong2022shadows} or a projector~\cite{lovisotto2021slap} to create shadows on targets as adversarial examples, aim for stealth but are vulnerable to changes in lighting conditions.

\subsection{Knowledge Distillation}
Knowledge distillation, a technique initially introduced by Hinton et al.~\cite{hinton2015distilling}, has emerged as a powerful approach for model compression and performance enhancement. By transferring knowledge from a larger, more complex teacher model to a smaller student model, knowledge distillation not only reduces the computational cost but also often improves the generalization capability of the student model. It has been widely applied to various tasks, including image classification~\cite{li2017learning}, object detection~\cite{li2017mimicking}, and image segmentation~\cite{mullapudi2019online}. More recently, knowledge distillation has also been explored in conjunction with adversarial learning~\cite{papernot2016distillation, papernot2017extending, goldblum2020adversarially, zi2021revisiting}. However, it serves as a defense mechanism, where adversarially trained teacher models guide student models to improve their robustness against adversarial attacks. 

Unlike using knowledge distillation for adversarial defense, we apply it to improve effectiveness of stealthy adversarial patches. Inspired by model distillation, we use a color-unconstrained patch as a teacher to guide the optimization of stealthy patches, ensuring both attack ability and invisibility. To our knowledge, it is the first time knowledge distillation has been applied to adversarial patch generation.

\section{Method}
\label{sec:methods}

\subsection{Problem Formulation}
\label{sec:Problem Formulation}
%In this work, we aim to optimize the adversarial patch within a color space matching the distribution of the surrounding environment.
Given a training dataset $\mathcal{D}(x,y)$, where $x$ represents clean image data and $y$ is the corresponding label, along with an object detector $f$, we introduce external adversarial knowledge $\Omega$ to facilitate the generation of a stealthy adversarial patch $\tau_s$ that effectively deceives the detector $f$. The optimization problem is formulated as follows:
\begin{equation}
\begin{split}
    \tau_s = \mathop{\arg\max}_{\tau_{s}\sim{\mathcal{T}_s}}\limits \mathbb{E}_{x\sim \mathcal{D}}\big[ \mathcal{L}(f,\mathcal{P}(x,\tau_s),y) + \\
    \Omega(f,\mathcal{P}(x,\tau_t),r) \big]
\label{fm:problem_definition}
\end{split}
\end{equation}
where $\mathcal{T}_s$ represents the target environment's color distribution, and \(\mathcal{P}(x, \tau)\) represents the operation of applying the adversarial patch \(\tau\) to the clean image \(x\), with the output serving as the adversarial example. Moreover, $\tau_t$ and $r$ represent teacher adversarial patch and related feature parameters, respectively. Additionally, $\mathcal{L}$ denotes a loss function that quantifies the discrepancy between the predictions of the adversarial example \(\mathcal{P}(x, \tau_{s})\) and the original label $y$.

% $\tau _{s}\sim \mathcal{T}_{s}$ and $\tau _{t}\sim \mathcal{T}_{t}$, $\mathcal{T}~$ represents the environmental distribution, Additionally, $\mathcal{L}$ denotes a loss function that quantifies the discrepancy between the predictions of the adversarial input $\sigma(x, \tau_s)$ and the original label $y$.

\subsection{Overall Framework}
\label{sec:Overall Framework}
As shown in Fig.~\ref{pic:framework}, our method comprises two key components: the stealthy adversarial patch generator and the adversarial knowledge distillation module. In the first component, we extract a color distribution that aligns with the target environment, which serves as the optimization space for the adversarial patch. We employ the reparameterization trick~\cite{maddison2014sampling} to optimize adversarial patch within this stealthy color space. In the second component, we transfer adversarial features from a color-unconstrained patch to enhance the attack effectiveness of the stealthy patch.

\subsection{Stealthy Adversarial Patch Generator}
\label{sec:Stealthy Adversarial Patch Generator}
\subsubsection{Stealthy Color Space}
To generate an adversarial patch that blends into the target environment, we first derive a color subset $\mathcal{C}$ that approximates the environment's color distribution $\mathcal{E}$. This objective can be formulated using the Kullback–Leibler divergence~\cite{csiszar1975divergence} between $\mathcal{C}$ and $\mathcal{E}$. We prioritize a more common set of colors from $\mathcal{E}$, ensuring better alignment with the environment’s color distribution.

The discrete environment color set $\mathcal{E} = {\epsilon_1, \epsilon_2, \dots, \epsilon_n}$ can be formed by collecting the values of each pixel in a target environment image. Given the desired number of color clusters $m$, our goal is to cluster these colors by minimizing the following objective.
\begin{equation}
\mathcal{C} = \mathop{\arg\min}_{c_1, c_2, \dots, c_m} \sum_{i=1}^{m} \sum_{\epsilon_j \in S_i} \|\epsilon_j - c_i\|^2
\end{equation}
where $S_i$ is the set of colors assigned to the $i$-th cluster, and $c_i$ is the center of the $i$-th cluster. Each color $\epsilon_j$ is assigned to the nearest cluster center: 
\begin{equation}
  \text{assign}(\epsilon_j) = \mathop{\arg\min}_{i} \|\epsilon_j - c_i\|
\end{equation}
The cluster centers $\mathcal{C} = {c_1, c_2, \dots, c_m}$ are iteratively updated until convergence, approximating the target environment's color distribution. To account for the differences between the RGB color model and human visual perception, we first convert the RGB color space to the LAB color space. Clustering is then performed in LAB space to identify a set of base colors, which are subsequently converted back to RGB, resulting in a stealthy color space for optimizing the adversarial patch.

\subsubsection{Differentiable Optimization in Stealthy Color Space}
Once the stealthy color set $\mathcal{C}$ is obtained, the next step is to constrain the optimization of the adversarial patch to this color set. We denote the color probability of the adversarial patch at position $(h,w)$ as $p(h,w)$. The color at position $(h,w)$ in the adversarial patch $\tau$ is then given by:
%  公式没有解释清楚
\begin{equation}
\begin{split}    
    \tau(h,w) = c_{k}, \text{where} \, k = \mathop{\arg\max}\limits_{i}{p_{i}(h,w)}
\end{split}
\label{fm:color_index}
\end{equation}

Since the $\arg\max$ operation is not differentiable, it cannot be directly optimized using gradient-based methods. To address it, we apply the reparameterization trick~\cite{maddison2014sampling}, which allows us to approximate the non-differentiable $\arg\max$ using a differentiable softmax function. Specifically, we independently sample from the $Gumbel(0,1)$ distribution to obtain $g_i$, and the color index $k$ at position $(h,w)$ can be expressed as:
\begin{align}
    k = \mathop{\arg\max}\limits_{i}{(g_i+\log{p_i(h,w)})} \label{fm:repa_k}
\end{align}
Next, we use a softmax estimator to approximate $\tau(h,w)$:
\begin{equation}
    \tau(h,w) = \sum\limits_{i=1}^{m}\frac{exp((g_i+\log{p_i(h,w)})/\omega)}{\sum_{j=1}^{m}{exp((g_j+\log{p_j(h,w)})/\omega)}}
\label{fm:repa_tau}
\end{equation}
where $\omega$ is the temperature coefficient. As $\omega$ decreases, the softmax distribution approaches a one-hot categorical distribution.

\subsection{Adversarial Knowledge Distillation Module} 
\label{sec:Object-Guided Distillation}
Adversarial patches optimized in an unconstrained color space usually exhibit stronger attack feature expressions. Drawing inspiration from model distillation, we introduce an adversarial knowledge distillation module, using a color-unconstrained patch as a teacher to guide the optimization of stealthy patches, enhancing attack performance. 

In our module, color-unconstrained adversarial examples are fed into the detection model to generate intermediate features $feat_{tch}$ with strong deception capabilities, while color-constrained adversarial examples produce features $feat_{stu}$, as illustrated in Fig.~\ref{pic:framework}. $feat_{tch}$ are used to guide the optimization of $feat_{stu}$, thereby improving the effectiveness of stealthy adversarial patches.

However, the features in $feat_{tch}$ and $feat_{stu}$ may contain irrelevant background information, posing a challenge for feature distillation. Therefore, we propose an adaptive feature weight mining mechanism that generates dynamic masks based on detection confidence scores or class scores from the output layer, filtered by a threshold $th$, Specifically, the target region $s(x, y)$ is defined as:
\begin{equation}
\begin{split}
    s_{n}(x, y) & = \max_{k \in \text{cls or obj}} s_{n,k}(x, y), \text{where } n \in \{t, s\}
\end{split}
\label{fm:interst region}
\end{equation}
where \( s_{n,k}(x, y) \) represents the score for the \( k \)-th class or detection confidence at position \( (x, y) \) of model \( n \), where \( n \) denotes either the teacher or student patch. Then the mask is computed as:
\begin{equation}
\begin{split}
    m_n(x, y) = \mathbb{I}(s_n(x, y) > th) \cdot s_n(x, y), \text{where } n \in \{t, s\}
\label{fm:msmt}
\end{split}
\end{equation}
where \(\mathbb{I}(\cdot > th)\) retains positions where the value exceeds the threshold $th$, setting others to zero. These regions are then mapped to the size of feature layer, generating masks \(\hat{m}_t\) and \(\hat{m}_s\) for the teacher and student patches. The difference between these masks defines the regions of interest \( M \), guiding optimization by highlighting areas requiring more attention.
\begin{align}
M(x, y) = |\hat{m}_t(x, y) - \hat{m}_s(x, y)|
\end{align}
Finally, the distillation objective \( \mathcal{L}_{distill} \) aims to transfer adversarial features from the teacher model to the student model, enhancing the attack effectiveness:
\begin{align}
    \mathcal{L}_{distill} = \sum\limits_{x,y}\Vert (feat_{tch}-feat_{stu})\cdot M \Vert_2  
\label{fm:loss_distill}
\end{align}

\begin{table}[t]
\caption{Adversarial patch attack performance on different detection models. The lower the value, the better the attack performance.}
\renewcommand{\arraystretch}{1.5}
\resizebox{\linewidth}{!}{%
\begin{tabular}{c|c|ccc|c}
\hline
\multirow{2}{*}{Method} & White Box & \multicolumn{3}{c|}{Black Box} & \multirow{2}{*}{Avg} \\ 
\cline{2-5} 
                        & YOLOv2   & YOLOv3 & YOLOv5 & FasterRCNN \\ \hline
Gray                    & 68.46     &  86.13  & 80.04   &  78.48 &  78.28   \\
Random                  & 68.30     &  86.47  &  81.99 &  85.02  &  80.45\\
White                   & 65.02     & 85.94  & 78.71   &  80.76  &  77.61   \\ \hline
NatPatch~\cite{hu2021naturalistic}                & 28.15     & 47.87   &  49.43  & \textbf{64.30}   &  47.44\\
DAP~\cite{guesmi2024dap}            & 27.74      & 52.60    &  49.24  & 68.61  & 49.55     \\ 
AdvCat~\cite{hu2023physically}         & 28.85     &  61.68  &  51.47  &  73.53 &  53.88   \\
Ours                    & \textbf{21.46}     & \textbf{45.46}   &  \textbf{41.24}  &  67.12 &  \textbf{43.82}     \\ \hline
\end{tabular}
}
\label{tab:compared}
\end{table}

\subsection{Optimization Objective}
\label{sec:Optimization Objective}
\textbf{Adversarial Loss.} The outputs of the student patch primarily contain the position and class of the object. Specifically, the predicted results for an input image $x$ are represented as $\hat{b}(x) = (\hat{b}_{pos}, \hat{b}_{cls}, \hat{b}_{obj})$, where $\hat{b}_{pos}$ denotes the object’s size and position, $\hat{b}_{cls}$ is the probability of the target class, and $\hat{b}_{obj}$ includes other relevant information such as foreground confidence. To deceive the model $f_s$ in terms of both class and position, the adversarial loss aims to decrease the probability of the target class and foreground confidence, while increasing the error in the predicted position. The adversarial loss function is defined as: 
\begin{align}
    \mathcal{L}_{adv} = \sum\limits_{x, y}\lambda_1\cdot{\hat{b}_{cls}}+\lambda_2\cdot{\hat{b}_{obj}}+\lambda_3\cdot{IoU(\hat{b}_{pos},{b}_{pos})}
    \label{fm:loss_adv}
\end{align}
where, $b(x)$ represents the ground truth for input $x$, and $IoU(\hat{b}_{pos}, b_{pos})$ is the intersection over union between the predicted $\hat{b}_{pos}$ and the ground truth $b_{pos}$.

The overall objective function $\mathcal{L}_{total}$ consists of two components: adversarial loss $\mathcal{L}_{adv}$ and distillation loss $\mathcal{L}_{distill}$:
\begin{align}
    \mathcal{L}_{total} = \mathcal{L}_{adv}+\beta\mathcal{L}_{distill}  \label{fm:loss_total}
\end{align}
where parameter $\beta$ is used to balance the contributions of different losses.

\section{Experiments}

\subsection{Experimental Setups}

\textbf{Datasets.} 
In digital experiments, we use the widely adopted INRIA dataset~\cite{dalal2005histograms}, comprising 614 training images and 288 test images with pedestrian bounding box annotations. For physical experiments, we test adversarial attacks in indoor and outdoor scenarios, using pedestrians with applied adversarial patches to validate the effectiveness of our proposed method.

\begin{figure}[tb] 
    \centering 
    \includegraphics[width=\columnwidth]{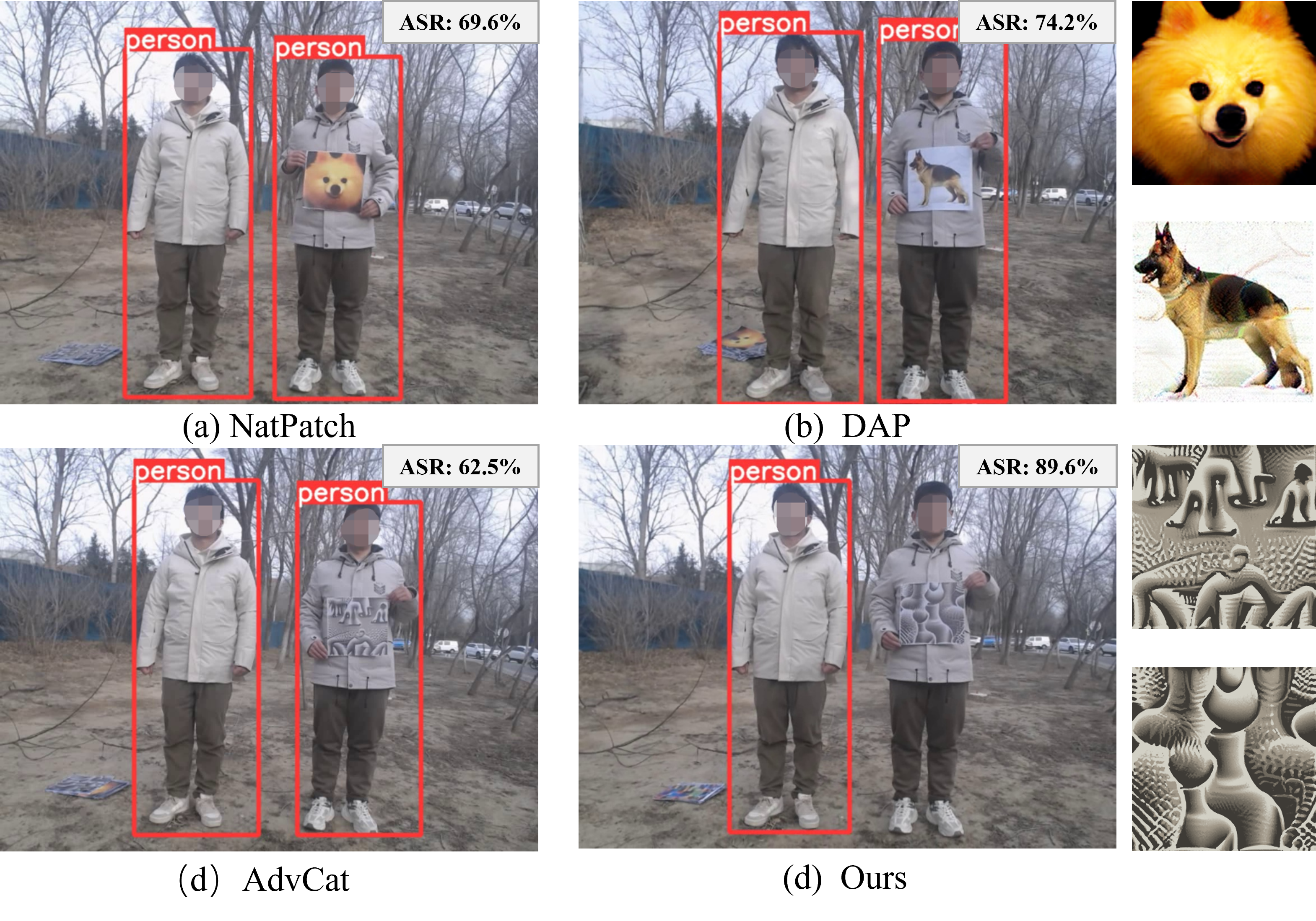}
    \caption{Compared with other SOTA methods in physical experiments. The top row compares other stealthy patch generation methods, namely NatPatch, DAP, and their corresponding adversarial patches from left to right. The bottom row compares the non-distillation-based method AdvCat, our approach, and the adversarial patches. Our method maintains better stealth while achieving stronger attack performance (ASR), effectively deceiving the YOLOv3 detector.}
    \label{pic:physical Experiments}
\end{figure}

\textbf{Baseline methods.}
To show the effectiveness of our proposed method, we compare it with six competitive algorithms in white-box and black-box settings: NatPatch~\cite{hu2021naturalistic}, AdvCat~\cite{hu2023physically}, DAP~\cite{guesmi2024dap}, and three common comparative patches (a gray patch, a random patch and a white patch, which we label as ``Gray", ``Random" and ``White"). It is worth noting that AdvCat is initially designed for optimization in 3D space, we have implemented its 2D patch version for a fair comparison.

\textbf{Optimization Details.}
In the experiment, the input image size is set to $640\times640$. Following previous work~\cite{thys2019fooling}, the size of the adversarial patch is set to $300\times300$, with a relative scaling ratio of 0.25 to the pedestrian bounding box. The batch size is 8, and the number of epoch is set to 300. In the EOT (Expectation over Transformation)~\cite{athalye2018synthesizing} steps, augmentation operations for the adversarial patch include random contrast, brightness, noise, and rotation. The parameter $\omega$ in formula \eqref{fm:repa_tau} is set to 0.3. For the distillation module, the two-stage model has a distillation coefficient \(\beta\) set to 0.01, while the single-stage model is set to 1. In selecting the feature layers, we selected the feature layer preceding the prediction layer for ease of indicating regions of interest using confidence scores. The filtering threshold of the feature layers $th$ is set to 0.25 for class scores and 0.1 for object confidence.

\begin{figure}[t] 
    \centering 
    \includegraphics[width=0.95\columnwidth]{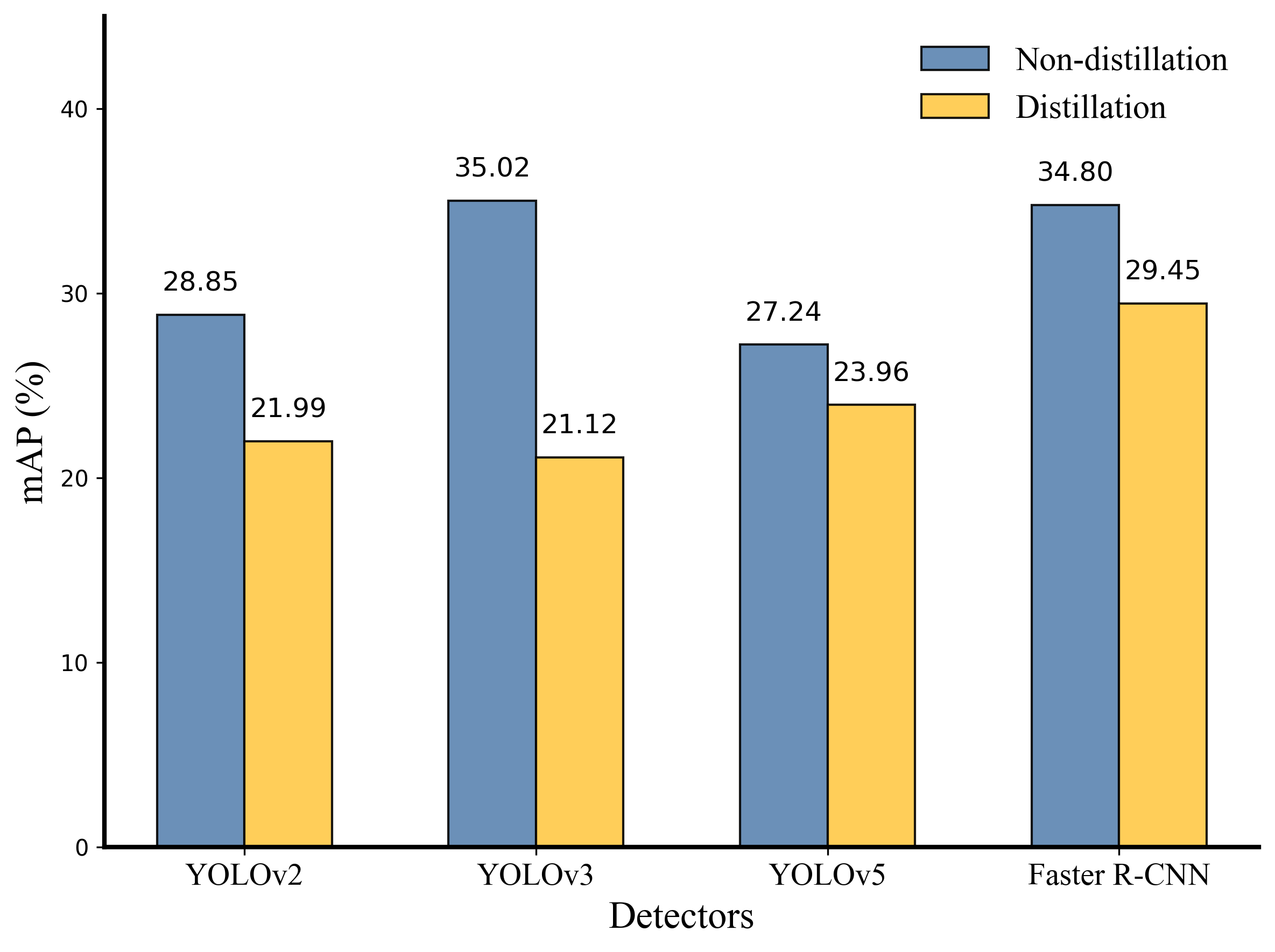}
    \caption{The attack performance with/without the distillation module. As shown, the distillation module leads to a significant drop in mAP on the INRIA test set.}
    \label{pic:diss_ablation}
\end{figure}

\textbf{Evaluation Metric.}
Following previous works~\cite{hu2021naturalistic,thys2019fooling,huang2023t}, we evaluate the effectiveness of our method in digital experiments using mean Average Precision at a 50 IoU threshold (mAP50). The evaluation metrics on the test set were obtained using the testing script provided in~\cite{huang2023t}, ensuring consistent experimental settings across different methods. For the physical-world experiments, we use the Attack Success Rate (ASR), defined as the ratio of successfully attacked frames to the total number of frames in a detection video.

\subsection{Digital Attack Experiments}
We use YOLOv2 as the white-box model and evaluate the effectiveness of different methods on three additional black-box models. We use gray, random noise, and white patches as the control group and compare our method with state-of-the-art (SOTA) approaches optimized in a constrained color space. Specifically, NatPatch and DAP use patches provided by their respective authors, while AdvCat, similar to our method, is optimized in the extracted color space and can be considered a generation method without distillation.

As shown in Table~\ref{tab:compared}, our method demonstrates significant advantages over the control group. Compared to other color-constrained patches, our method shows a notable improvement in attack performance on white-box models, with mAP50 reductions exceeding $20\%$. The average mAP of NatPatch, DAP, and our method on three black-box models are 53.87, 56.82, and 51.27, respectively, with our method showing an average improvement of approximately $7\%$. Compared to the non-distillation method AdvCat, our approach achieves more than a $20\%$ improvement in attack performance on both white-box and black-box models, further validating the effectiveness of our distillation framework. The visual results of the digital experiments are presented in the supplementary materials.

\subsection{Physical Attack Experiments}
We have conducted physical-world experiments on typical detection models.  Firstly, the dominant colors of the environment has been extracted from real-world scenes to construct a concealed color set. As shown in Fig.~\ref{pic:physical Experiments}, we compared our method with other competitive methods in physical environments. The results demonstrate that our adversarial patches more effectively deceive the detector. Furthermore, from a visual perspective, the generated patches blend seamlessly with the surrounding environment, achieving superior stealthiness.

\begin{figure}[tb] 
    \centering 
    \includegraphics[width=0.95\columnwidth]{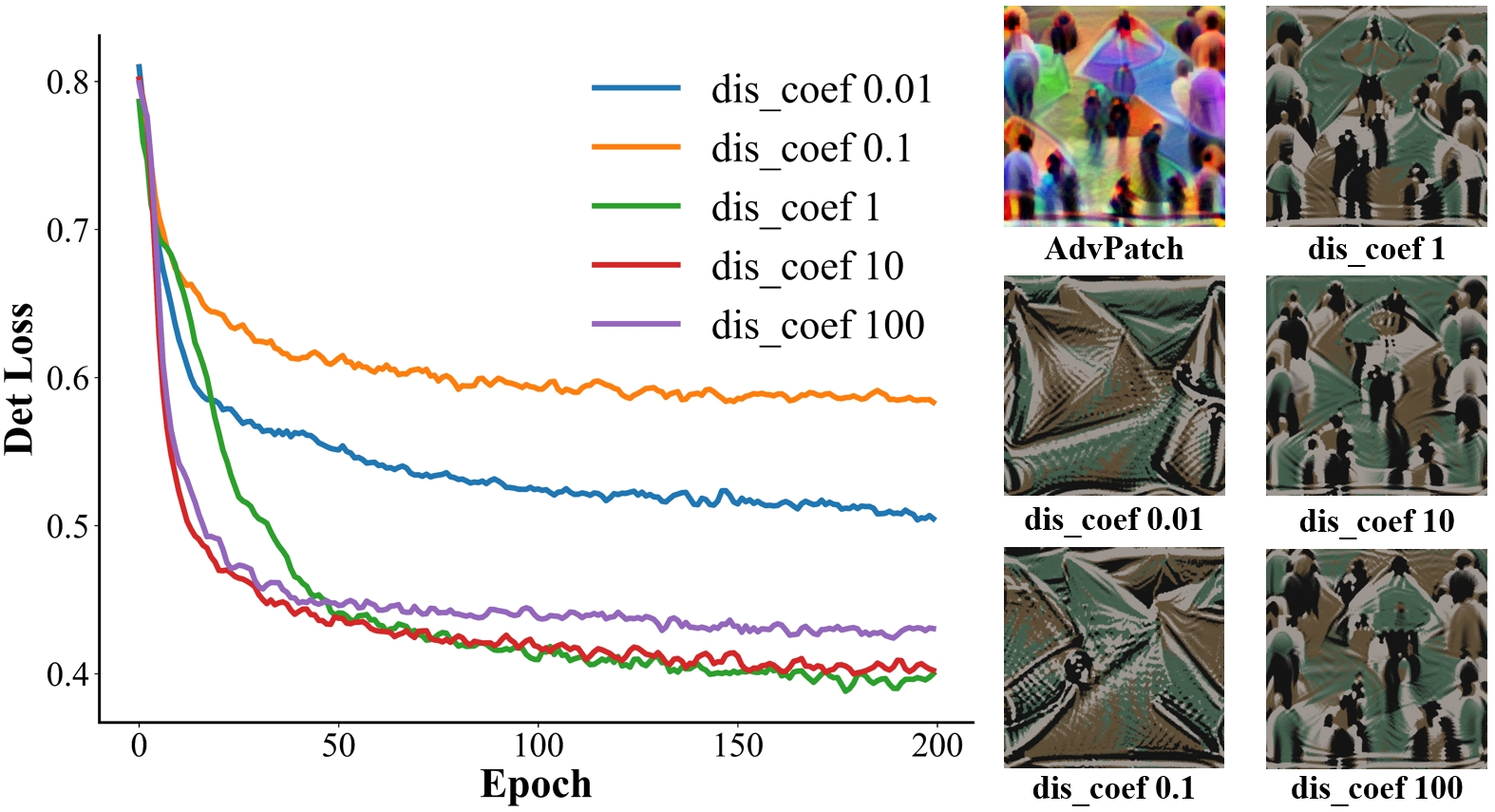}
    \caption{
The YOLOv5 decline curve of detection box confidence under different distillation loss coefficients.}
    \label{pic:coef of dis loss}
\end{figure}

\subsection{Ablation Study}
To validate the effectiveness of the proposed distillation module, we conducted quantitative evaluation experiments on typical detection models. We compared the attack performance of adversarial patches optimized in the concealed color space, both without the distillation method~\cite{hu2023physically} and with the distillation method. As illustrated in the Fig.~\ref{pic:diss_ablation}, it is evident that our distillation-based method improves attack performance by over $20\%$ while maintaining stealthy.

\subsection{Discussions}
\textbf{Impact of the Distillation Loss Coefficient.}
In experiments, we observed that different distillation loss coefficients may impact on knowledge transfer. Here, we explore how distillation coefficients affect the detection confidence scores. As illustrated in Fig.~\ref{pic:coef of dis loss}, a higher distillation coefficient can effectively reduce the gap between the teacher patch (AdvPatch~\cite{thys2019fooling}) and the student patch, making the patterns increasingly close. This indirectly validates the effectiveness of the distillation module in our proposed method. However, excessively high distillation coefficients make further optimization challenging and lead to performance degradation. Based on the experimental results, we set the distillation loss coefficient 
 $\beta$ to 1.

\textbf{Feature Mask $M$ Strategy Used in Distillation.}
The predicted detection result \(P\) includes the bounding box coordinates, detection confidence \(obj\_conf\), and the maximum class score \(cls\_max\_conf\). Based on \(obj\_conf\) and \(cls\_max\_conf\), we can derive masks that indicate the important regions of the distilled features layers. We explore four strategies for mapping feature importance: \(obj\_conf\), \(cls\_max\_conf\) , \(obj\_conf \mid cls\_max\_conf\), and \(obj\_conf ~\&~cls\_max\_conf\). According to formula \eqref{fm:interst region} and \eqref{fm:msmt}, the confidence scores are used to determine whether a region should be included in the calculation of the distilled loss for optimization. As shown in the Table ~\ref{tab:mask strategy}, both single-stage and two-stage detectors achieve favorable results when using \(cls\_max\_conf\) to filter features. Due to differences in candidate box extraction methods, the two-stage detector performs less effectively when using proposal confidence \(obj\_conf\) for feature filtering.

\begin{table}[t]
\caption{The Impact of Different Mask Processing Strategies on detection performance. The lower the value, the better the attack performance.}
\resizebox{1.0\columnwidth}{!}{
\renewcommand{\arraystretch}{1.3}
\centering
\begin{tabular}{c|c|c}
\hline
Mask Strategy               & YOLOv2  & FasterRCNN \\ \hline
 \(obj\_conf\)            &    21.13  & 59.51 \\ \hline
\(cls\_max\_conf\)        &     \textbf{20.80}  &  31.14 \\ \hline
\(obj\_conf \mid cls\_max\_conf\)    & 27.55  &   \textbf{29.45}\\ \hline
\(obj\_conf~\&~cls\_max\_conf\)    &26.22   &   53.04\\ \hline
\end{tabular}
\label{tab:mask strategy}
}
\end{table}

\section{Conclusion}
In this paper, we propose a novel physical attack method utilizing knowledge distillation. By incorporating a primary color extractor, we constrain the adversarial patch optimization to a color space that aligns with the surrounding environment. The distillation framework employs an effective teacher patch in adversarial attack to guide the creation of stealthy adversarial patches. Through extensive experiments, the proposed method improve attack performance about 20\%, offering enhanced effectiveness compared to existing state-of-the-art methods, while maitaining a high level of stealth.

\bibliographystyle{IEEEbib}
\bibliography{icme2025references}

%\clearpage
%\input{sec/6_appendix}
\end{document}